%% file: iclr2026_conference.tex
\documentclass{article} 
\usepackage{iclr2026_conference,times}
\iclrfinalcopy
\input{math_commands.tex}

\usepackage{hyperref}
\usepackage{url}
\usepackage{algorithm}
\usepackage{algorithmic}
\usepackage{graphicx} 
\usepackage{caption}
\usepackage{subcaption}
\usepackage{multirow}
\usepackage{makecell}
\usepackage{pifont}
\newcommand{\cmark}{\ding{51}}
\newcommand{\xmark}{\ding{55}}
\usepackage{xcolor}
\makeatletter
\DeclareRobustCommand\onedot{\futurelet\@let@token\@onedot}
\def\@onedot{\ifx\@let@token.\else.\null\fi\xspace}
\usepackage{booktabs}
\usepackage{amsmath}
\usepackage{amssymb}

\makeatother
\usepackage[capitalize]{cleveref}
\crefname{section}{Sec.}{Secs.}
\Crefname{section}{Section}{Sections}
\Crefname{table}{Table}{Tables}
\crefname{table}{Tab.}{Tabs.}
\title{Enhancing Open-Vocabulary Object Detection through Multi-Level Fine-Grained Visual-Language Alignment}

\author{Tianyi Zhang \thanks{Work done while the author was an intern at Meta.} \\
University of Minnesota\\
\texttt{\{zhan9167\}@umn.edu} \\
\And
Antoine Simoulin \& Kai Li \\
Meta \\
\texttt{\{antoinesimoulin,kailee88\}@meta.com} \\
\AND
Sana Lakdawala \& Shiqing Yu \& Arpit Mittal \& Hongyu Fu \\
Meta \\
\texttt{\{sanalakdawala,sqy,arpitmittal,hongyufu\}@meta.com} \\
\AND
Yu Lin \\
Meta \\
\texttt{\{yulin0077\}@meta.com} \\
}

\begin{document}

\maketitle

\begin{abstract}
Traditional object detection systems are typically constrained to predefined categories, limiting their applicability in dynamic environments.
In contrast, open-vocabulary object detection (OVD) enables the identification of objects from novel classes not present in the training set.
Recent advances in visual-language modeling have led to significant progress of OVD.
However, prior works face challenges in either adapting the single-scale image backbone from CLIP to the detection framework or ensuring robust visual-language alignment.
We propose Visual-Language Detection (VLDet), a novel framework that revamps feature pyramid for fine-grained visual-language alignment, leading to improved OVD performance.
With the VL-PUB module, VLDet effectively exploits the visual-language knowledge from CLIP and adapts the backbone for object detection through feature pyramid.
In addition, we introduce the SigRPN block, which incorporates a sigmoid-based anchor-text contrastive alignment loss to improve detection of novel categories.
Through extensive experiments, our approach achieves 58.7 AP for novel classes on COCO2017 and 24.8 AP on LVIS, surpassing all state-of-the-art methods and achieving significant improvements of 27.6\% and 6.9\%, respectively. 
Furthermore, VLDet also demonstrates superior zero-shot performance on closed-set object detection.
\end{abstract}

\section{Introduction}
\label{sec:intro}

With the development of visual-language models (VLM) \citep{radford2021learning, zeng2023x, zhai2023sigmoid}, recent approaches have extended traditional object detection frameworks into open-world object detectors capable of detecting objects from categories not present in the training set \citep{li2022grounded, zhong2022regionclip, liu2023grounding, cheng2024yolo, zhou2022detecting, gao2022open}. 
As a pioneer, OVR-CNN \citep{zareian2021open} proposed open-vocabulary object detection by introducing a pre-training stage that uses image-caption pairs to align image and text encoders.
Then, the detection model employs the aligned image encoder as its backbone, classifying bounding boxes based on the similarity between image features and text embeddings of category names generated by the text encoder.
Although subsequent works \citep{gu2021open, wu2023aligning} have improved the performance by distilling knowledge from CLIP \citep{radford2021learning} for region-wise classification, they still rely on visual-language knowledge aligned on image-level.
A fine-grained region-wise visual-language alignment is desired for more accurate detection \citep{zhong2022regionclip}.
Many works with region-wise alignment have been proposed to bridge this gap, such as RegionCLIP \citep{zhong2022regionclip} 
and YOLO-World \citep{cheng2024yolo}. 
However, these approaches also bring several challenges. 
First, they struggle to adapt image encoders from CLIP into detection models. 
Some methods simply adopt the single-scale backbone from CLIP, which compromises spatial information and detection accuracy.
Others attempt to train a multi-scale backbone from scratch using pseudo-labeled regions (\emph{i.e.}, bounding boxes) generated by zero-shot detectors, such as GLIP \citep{li2022grounded}, on large-scale datasets.
These pseudo labels demand significant additional processing efforts \citep{li2022grounded, yao2022detclip, liu2023grounding, cheng2024yolo}, which greatly limits their reproducibility. 
\begin{figure}
  \centering
  \includegraphics[width=0.9\columnwidth]{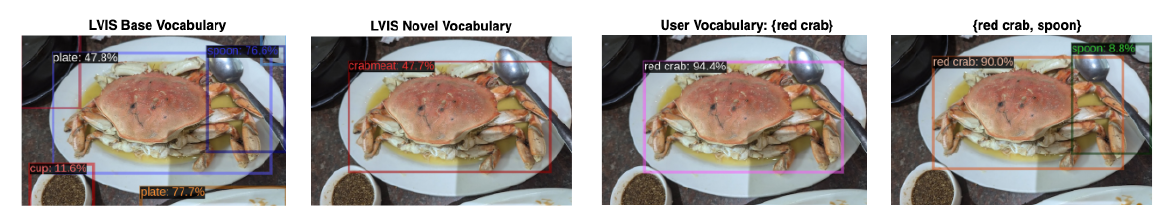}
  \caption{With multi-level fine-grained visual-language alignment, VLDet accurately detects objects from LVIS categories to user-specified ones. }
  \vspace{-5mm}
  \label{fig:intro}
\end{figure}
Second, these methods bias toward region-wise alignment, ignoring that image-wise alignment is also beneficial for visual-language modeling \citep{zeng2023x}.
Finally, aforementioned methods either adopt a single-stage detector, or leverage a vanilla Region Proposal Network (RPN).
A sophisticated RPN can significantly boost OVD performance by proposing objects of any potential classes \citep{gu2021open}.

To tackle these challenges, we propose a novel framework, Visual-Language Detection (VLDet).
By introducing a Visual-Language Pyramid Upscale Block (VL-PUB), we integrate CLIP's image and text encoders into our framework to effectively exploit its enriched visual-language latent space and construct a feature pyramid to detect objects of varying sizes.
To enhance the visual-language alignment, we incorporate contrastive losses at multiple levels, achieving multi-level fine-grained alignment. Specifically, we employ: (i) an image-caption contrastive loss, which provides a broader understanding of the relationship between full images and captions. This is conducted within mini-batches to balance training between image-wise and region-wise visual-language alignment; (ii) a region-text contrastive loss \citep{li2022grounded} to achieve fine-grained alignment between image regions and object category texts; (iii) an additional anchor-text binary visual-language alignment, which enhances the RPN ability to differentiate background from objects of any category and facilitate learning general semantic information applicable to any category, thereby improving generalizability to novel categories, as illustrated in \cref{fig:intro}.
Notably, VLDet is pre-trained solely on the detection dataset, Objects365 \citep{shao2019objects365}, where arbitrary foundation VLM can be leveraged to generate captions. 
This approach offers significant efficiency and reproducibility compared to the pseudo-labeling of bounding boxes for large-scale image-caption pairs.
Through comprehensive experiments on public datasets for OVD, VLDet outperforms all baselines and also demonstrates extraordinary performance of zero-shot inference and fine-tuning for traditional object detection.

To summarize, our main contributions are as follows,
\begin{itemize}
    \item We propose a VL-PUB module to effectively adapt image and text encoders from CLIP into open-world object detector to leverage the enriched visual-language latent space and utilize multi-scale image features for enhanced detection performance.
    \item We design SigRPN, the first RPN for OVD that uses sigmoid-based visual-language contrastive learning to differentiate between background and objects of any class, aiding in proposing objects of all potential new categories and significantly enhancing generalization to novel categories in the OVD scenario. We further integrate image-wise contrastive loss for more robust visual-language alignment.
    \item Through extensive experiments on public datasets, the proposed VLDet outperforms state-of-the-art methods in OVD, particularly enhancing the mAP of novel categories for OVD from 46.0 to 58.7 on COCO2017, and from 23.2 to 24.8 on LVIS, showing the great generalization capability of our VLDet.
\end{itemize}

\section{Related Work}
\begin{figure*}
  \centering
  \includegraphics[width=0.9\textwidth]{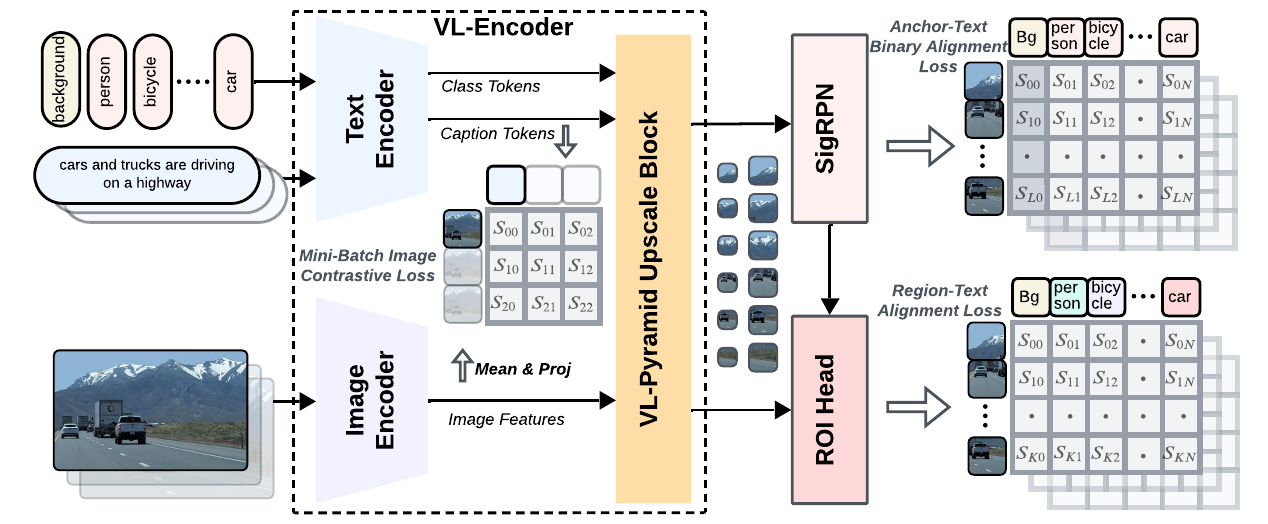}
  \caption{With multi-level visual-language alignment, VLDet is a unified network which can effectively exploit enriched visual-language semantic space from pre-aligned single-scale backbones while also providing pyramid image features for improved performance in OVD.}
  \vspace{-6mm}
  \label{fig:method}
\end{figure*}
\label{sec:related work}

\subsection{Traditional Object Detection}
Traditional object detectors are trained on datasets with predefined categories and subsequently detect objects within this fixed set of categories (\emph{i.e.}, Closed-set). Popular object detectors can be categorized into three groups: single-stage, two-stage, and query-based. 
YOLOs \citep{redmon2016you, redmon2017yolo9000, redmon2018yolov3, yolo11_ultralytics} are the representative works from the first category, which utilize convolutional architectures for real-time object detection. 
DETR \citep{carion2020end} pioneered the use of transformers \citep{zhang2024context, cai2025does, zhang2024transformer, zhang2024patch} for object detection, inspiring numerous query-based methods \citep{kamath2021mdetr, zhang2022dino, zong2023detrs, ren2023strong}. Two-stage detectors \citep{ren2016faster, he2017mask, cai2019cascade}, such as Faster R-CNN \citep{ren2016faster} employ a two-stage framework for proposal generation and Region-of-Interest (RoI) classification and regression. 
While RPN in two-stage learns general information to distinguish background from objects across all categories, whether novel or base, this paper emphasizes the importance of RPN for OVD.

\subsection{Open-world Object Detection}
There are two primary branches in the realm of open-set object detection: open-vocabulary object detection (OVD) and zero-shot object detection (ZOD). 

OVD was first proposed by OVR-CNN \citep{zareian2021open}, where a visual encoder is pre-trained on image-text pairs to learn object concepts, enabling the detection of additional categories not included in the training set. Subsequently, ViLD \citep{gu2021open} and DetPro~\citep{du2022learning} trained object detectors by distilling visual features from a pre-trained CLIP model. RegionCLIP \citep{zhong2022regionclip} further advanced this by learning region representations from "pseudo" region-text pairs provided by a pre-trained CLIP model.  Detic \citep{zhou2022detecting} trained the classifier of a detector on ImageNet21K with broader vocabularies.
BARON \citep{bica2024improving} groups contextually interrelated regions as a bag and aligns the embedding of the bag of regions beyond individual regions.

In the context of ZOD, recent methodologies, beginning with GLIP \citep{li2022grounded, liu2023grounding, shen2024aligning, cheng2024yolo, jiang2024t}, achieve it by integrating the phrase grounding task with object detection.
This integration leverages grounding datasets (\emph{e.g.} GoldG curated by \citep{kamath2021mdetr} from \citep{plummer2015flickr30k, krishna2017visual, hudson2019gqa}) and the large-scale image-caption pairs \citep{sharma2018conceptual, ordonez2011im2text} with pseudo-labeled bounding boxes, which offer a multitude of categories by extracting words from image captions. 
Our approach primarily targets OVD, yet it also demonstrates promising ZOD capability, achieved by pre-training solely on Objects365 and avoiding bounding box pseudo-labeling, thereby enhancing reproducibility.

\section{Approach}
\label{sec:approach}
Single-scale ViT backbones are widely adopted by researchers for visual-language alignment with image-caption pairs \citep{radford2021learning, zareian2021open, zhai2023sigmoid}. They produce enriched multi-modal embedding space but yield limited spatial information. 
To address this, VLDet is designed to unify any single-scale visual-language backbones into object detectors, providing features at various scales for enhanced detection, as illustrated in \cref{fig:method}.

\subsection{Overall Architecture} \label{sec:overall_arch}
To leverage the enriched semantic information from CLIP's aligned image and text encoders, we design a Visual-Language Encoder (VL-Encoder) that integrates these two encoders with a Visual-Language Pyramid Upscale Block (VL-PUB). While the CLIP image encoder extracts features at a single scale, VL-PUB casts the output features into multiple scales after fusing them with text features, enhancing detection performance for objects of different sizes \citep{lin2017feature}.
Previous work \citep{li2022grounded} grouped all class names together and generated embeddings for this large sequence, which led to two significant issues: first, it assigns a different number of tokens to different classes, requiring token grouping when computing similarities with visual features;
second, nontrivial adjustment is required when the dataset (\emph{e.g.} Object365 or LVIS) contains too many classes, which exceeds the maximum token length of the text encoder (\emph{e.g.} 256);
In contrast, we embed each category name into one single token using the CLIP text encoder, which is more systematic for contrastive loss computation and naturally supports an arbitrary number of categories without additional modifications. 
Moreover, it's worth noting that we only use class names as text input during inference, while an extra image caption branch is included at training stage.
This design benefits text feature extraction for class names by providing more textual context during training.
By leveraging such an informative text feature when conducting image-wise contrastive learning, we further enhance the visual-language alignment for object detection.

Specifically, an input image, $\boldsymbol{x} \in \mathbb{R}^{H\times W\times 3}$ ($H$ is image height, and $W$ is the width), first goes through the image encoder and is embedded into a single-scale image feature, $\boldsymbol{v}_0 \in \mathbb{R}^{\frac{H}{p}\times\frac{W}{p}\times C_v}$, where $C_v$ is the number of visual dimensions and $p$ is patch size. Class prompts are embedded into language feature, $\boldsymbol{l}_{cls}\in\mathbb{R}^{N\times C_l}$, where $N$ is the number of classes in the training data including \textit{background} class (\emph{e.g.} 366 for Objects 365) and $C_l$ is the dimensions of language feature. 
The image caption is also embedded by text encoder into a single token, $\boldsymbol{l}_{cap}\in \mathbb{R}^{1\times C_l}$. 
We add a linear layer to project the mean of extracted image feature $\boldsymbol{v}_0$ over all visual tokens to the same dimension of caption embedding $\boldsymbol{l}_{cap}$ for image-level visual-language contrastive learning, leading to better visual-language alignment. 

To better capture the objects of various sizes, the extracted text features $\boldsymbol{l}_{cls}, \boldsymbol{l}_{cap}$ and image feature $\boldsymbol{v}_0$ first go through VL-PUB to generate image features at multiple scales. This block first fuses the image feature with text feature via bi-directional cross-attention \citep{li2022grounded}, then employ a single transformer layer to further embed text features into deeper embeddings $\boldsymbol{l}'_{cls}\in \mathbb{R}^{N\times C_l}$ and leverage multiple Deconvolution, MaxPooling and Convolution layers to generate image feature at various scales $\boldsymbol{v}_i$, where $i\in\{1, 2, ..., Z\}$ and $Z$ is a hyper-parameter, denoting the number of output scales.
The scales of generated image features range from $\frac{4H}{p}\times\frac{4W}{p}$ to $\frac{H}{4p}\times\frac{W}{4p}$, and the visual dimension $C_v$ becomes 256. 
For simplicity, we use $C_v=256$ in the following sections.

The extracted multi-scale image features $\boldsymbol{v}_i, i\in\{1, 2, ..., Z\}$ and text feature $\boldsymbol{l}'_{cls}$ are fed into an Visual-Language RPN module (SigRPN) for proposal prediction of any potential object classes. 
In SigRPN, the text feature is first fused with multi-scale image features through bi-directional cross-attention \citep{li2022grounded} then goes through a single transformer layer to get a more informative feature $\boldsymbol{l}''_{cls}\in\mathbb{R}^{N\times C_l}$. After the fusion with text feature, the image features with various scales are fed into several convoluation layers. Each scale $\boldsymbol{v}_i\in\mathbb{R}^{h_i\times w_i\times C_v}$ ($h_i$ and $w_i$ denotes the height and width of the feature map for the $i$th scale) generates the corresponding \textit{objectness logits} $\boldsymbol{a}_i\in \mathbb{R}^{h_i\times w_i \times A \times C_l}$, where $A$ denotes the number of anchors for each spatial position and \textit{anchor deltas} $\boldsymbol{d}_i\in \mathbb{R}^{h_i\times w_i \times A \times 4}$ \citep{ren2016faster}. 
Finally, based on the proposals from SigRPN, ROI Head extracts region-wise features from the multi-scale image features $\boldsymbol{v}_i, i\in\{1, 2, ..., Z\}$ for object classification ($Z=5$ in our experiment).


\subsection{Visual-Language Pyramid Upscale Block} \label{sec:vl-pub}
\begin{figure}
  \centering
  \includegraphics[width=0.85\columnwidth]{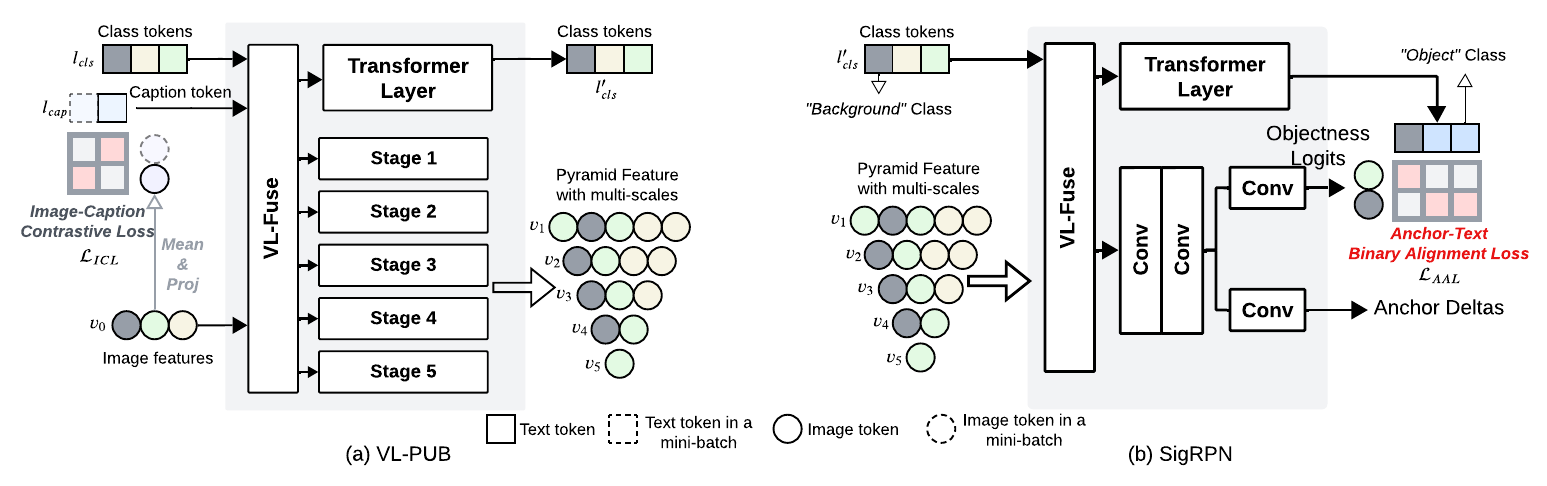}
  \caption{(a) $\mathcal{L}_{ICL}$ for image-wise visual-language alignment is computed before VL-PUB module. After the fusion of image feature and text feature, VL-PUB generates deeper text feature and pyramid image feature at various scales. (b) SigRPN computes $\mathcal{L}_{AAL}$ for fine-grained visual-language alignment on the general difference of \textit{Background} and \textit{Object} of any class.}
  \vspace{-4mm}
  \label{fig:vlrpn}
\end{figure}

Although the pre-aligned image and text encoders from CLIP contain enriched semantic information, having been trained on a dataset with 400M image-caption pairs, the image encoder outputs visual features at only a single scale. 
Multi-scale image features typically perform much better \citep{lin2017feature, li2022exploring} for object detection, as objects in images vary in size, and multi-scale feature maps can provide more comprehensive spatial information.
Follwing the same idea, before feeding the image features generated by the image encoder directly into the subsequent RPN and ROI Head, we introduce a Visual-Language Pyramid Upscale Block (VL-PUB) to extract feature maps at various scales, as shown in \cref{fig:vlrpn}.

Specifically, suppose we have obtained the text features $\boldsymbol{l}_{cls}$, $\boldsymbol{l}_{cap}$ and image feature $\boldsymbol{v}_0$ from text encoder $\text{E}_L$ and image encoder $\text{E}_V$ as follows,
\begin{equation}
    \boldsymbol{l}_{cls} = \text{E}_L(\text{CLS}), \;\; \boldsymbol{l}_{cap} = \text{E}_L(\text{CAP}),\;\; \boldsymbol{v}_0 = \text{E}_V(\boldsymbol{x}),
\end{equation}
where $\boldsymbol{x}$ is the input image, CAP, CLS denotes the corresponding caption and class prompts. We encode CLS and CAP separately to reduce the memory consumption and computation by reducing the number of tokens for CLS since the category names are much shorter than captions.

Previous works proved that early fusion of visual and language features leads to enhanced performance \citep{li2022grounded, liu2023grounding}. We first employed an visual-language fuse layer (VL-Fuse) to enhance both text and image features with bi-directional cross-attention (Detailed in \cref{sec:crossattn}) after concatenating the class tokens and caption token as follows,
\begin{equation} \label{eq:single}
    \boldsymbol{v}_f, \boldsymbol{l}_f = \text{VLFuse}(\boldsymbol{v}_0, \text{Concat}(\boldsymbol{l}_{cls}, \boldsymbol{l}_{cap}))
\end{equation}
where $\boldsymbol{v}_f$, $\boldsymbol{l}_f$ denotes the fused visual feature and language feature. Next, the text feature passes through a single transformer layer to generate deeper text embeddings $\boldsymbol{l}'_{cls}$.

For the image feature, VL-PUB employs a pyramid feature module with multi-stages to generate image feature with various scales after the fusion with text embeddings. 
Following \cite{lin2017feature, li2022exploring}, we leveraged Deconvolution layers and MaxPooling layers for the upscaling and downscaling of feature maps, resulting in feature maps with 5 different scales.

\subsection{Visual-Language Region Proposal Network} \label{sec:vl-rpn}
Since RPN in OD aims to propose regions for any class, whether base or novel, we emphasize the importance of the RPN module for OVD in this section.
Given previous success of region-based visual-language alignment for region classification \citep{zhong2022regionclip, li2022grounded} and the superority of sigmoid loss over softmax contrastive loss \citep{zhai2023sigmoid}, we propose a novel SigRPN block with sigmoid-based visual-language alignment to enhance the model’s generalization capability to novel classes.

Specifically, a VL-Fuse layer first fuses the text feature with all visual features of various scales using bi-directional cross-attention to enhance context understanding and generate more informative feature representations as follows,
\begin{equation}\label{eq:multi}
    \boldsymbol{v}'_f, \boldsymbol{l}'_f = \text{VLFuse}( \text{Concat}(\boldsymbol{v}_1, \boldsymbol{v}_2, \boldsymbol{v}_3, \boldsymbol{v}_4, \boldsymbol{v}_5), \boldsymbol{l}'_{cls})
\end{equation}
where $\boldsymbol{v}'_f$, $\boldsymbol{l}'_f$ denotes the fused visual feature and language feature separately after the VL-Fuse layer of SigRPN. And the text feature will go through a single transformer layer and generate more contextual embeddings $\boldsymbol{l}''_{cls}$.

For the visual feature maps, two convolution layers first enhance the visual representation for each scale. 
We then fed the enhanced visual features to two parallel convolution layers to generate \textit{objectness logits} and \textit{anchor deltas} per anchor.
\textit{objectness logits} are used for proposal binary classification to determine whether this anchor belongs to background or foreground objects in a contrastive way.
\textit{anchor deltas} are used to transform anchors into box proposals. 


\subsection{Multi-level Fine-grained Alignment Loss} \label{sec:loss}
To enable more robust visual-language alignment, we introduce alignment loss into multiple stages and effectively combine them to achieve better performance.

\textbf{Mini-Batch Image Contrastive Loss.}
To fully exploit the pre-aligned image and text encoders from CLIP, we absorb the conventional CLIP training loss into our VLDet framework for image-wise alignment.
Although CLIP recommends a large batch size for contrastive learning, we found it results in worse detection performance since it leads to severe loss balance issues and emphasis too much on the image-wise visual-language alignment compared with RPN and ROI losses. 
To reduce the scale of image-wise contrastive loss and prevent it from dominating the training process, we divide inputs into mini-batches and compute the image-wise contrastive loss as follows,
\begin{equation}
        \mathcal{L}_{ICL} = -\frac{1}{2B}\sum_{k=1}^{B/M}\sum_{m_k=1}^M(log\frac{exp(\phi(\boldsymbol{v}_{m_k}, \boldsymbol{l}_{m_k})/\tau)}{\sum_{n_k=1}^Mexp(\phi(\boldsymbol{v}_{m_k}, \boldsymbol{l}_{n_k})/\tau)}\\
        +log\frac{exp(\phi(\boldsymbol{l}_{m_k}, \boldsymbol{v}_{m_k})/\tau)}{\sum_{n_k=1}^Mexp(\phi(\boldsymbol{l}_{m_k}, \boldsymbol{v}_{n_k})/\tau)})
\end{equation}
where $B$ is the batch size, and $M$ is a hyper-parameter denoting the mini-batch size.
During our experiment, we found $M=8$ yields the best trade-off between the losses. $\phi(\boldsymbol{v}, \boldsymbol{l})$ is the cosine similarity between visual and language embeddings and $\tau$ is the temperature \citep{wu2018unsupervised}. 



\textbf{Anchor-Text Binary Alignment Loss.}
Besides the image-wise contrastive loss that aligns the image feature and text feature in a coarse-level, we introduce two region-wise alignment losses for fine-grained alignment.
The first one is Anchor-text binary Alignment Loss (AAL) within the SigRPN module.
Conventional RPNs typically solve the background-vs-foreground problem via a binary classification.
In SigRPN, we tackle this problem in a constrastive manner.
Specifically, we first compute the similarities of visual features with text features of all classes, including \textit{background}.
We then use the mean of similarities over all object classes to subtract the similarity with \textit{background} as the final \textit{objectness score} $\boldsymbol{s}_{obj}$ as follows,
\begin{equation}
    \boldsymbol{s}_{obj} = (\frac{1}{N'}\sum_{i=1}^{N'}\phi(\boldsymbol{v}, \boldsymbol{l}_i) - \phi(\boldsymbol{v}, \boldsymbol{l}_0))/\tau
\end{equation}
where $N'$ denotes the number of foreground object categories, and $\boldsymbol{l}_0$ is the text feature of \textit{background}. We wrap $\boldsymbol{s}_{obj}$ within a binary cross entropy (BCE) loss as the RPN classification loss $\mathcal{L}_{AAL}$.

\textbf{Region-Text Alignment Loss.}
For ROI, we wrap the similarity of visual features and text features into categorical cross entropy as region-text alignment loss $\mathcal{L}_{RAL}$  by following \cite{li2022grounded}.


\begin{table*}
  \centering
  \caption{VLDet outperforms all baselines for $Novel$ categories for OVD on COCO2017 and LVIS.}
  \resizebox{\textwidth}{!}{%
  \begin{tabular}{@{}l l l |c c c | c c c c@{}}
    \toprule
    \multirow{2}{*}{\textbf{Method}} & \multirow{2}{*}{\textbf{Pre-train Data}} & \multirow{2}{*}{\textbf{Backbone}} & \multicolumn{3}{|c}{\textbf{COCO2017}} & \multicolumn{4}{|c}{\textbf{LVIS}}\\
    & & & $\text{AP}_{Novel}$ & $\text{AP}_{Base}$ & $\text{AP}_{All}$ & $\text{AP}_{r}$ & $\text{AP}_{c}$ & $\text{AP}_{f}$ & AP\\
    \midrule
    OVR-CNN~\citep{zareian2021open} & COCO Caption & RN50 & 22.8 & 46.0 & 39.9 & - & - &- &- \\
    ViLD~\citep{gu2021open} & CC3M & RN50 & 27.6 & \textbf{59.5} & 51.3 & 16.7 & 26.5 & 34.2 & 27.8\\
    RegionCLIP~\citep{zhong2022regionclip} & CC3M & RN50 & 31.4 & 57.1 & 50.4 & 17.1 & 27.4 & 34.0 & 28.2\\
    F-VLM~\citep{kuo2022f} & CLIP & RN50 & 28.0 & - & 39.6 & 18.6 & - & - & 24.2\\
    
    BARON~\citep{wu2023aligning} & CLIP+COCO Cap & RN50 & 42.7 & 54.9 & 51.0 & 23.2 & 29.3 & 32.5 & 29.5 \\
    
    YOLO-World-M~\citep{cheng2024yolo} & O365+Gold & Y8-M & - & - & - & 15.9 & 24.6 & 39.0 & 28.8\\
    YOLO-World-L~\citep{cheng2024yolo} & O365+Gold+CC3M & Y8-L & - & - & - & 20.4 & 31.1 & 43.5 & 34.1\\
    OV-DQUO~\citep{wang2025ov} & CLIP & RN50x4 & 45.6 & - & 48.1 & - & - & - & - \\
    CCKT-Det++~\citep{zhang2025cyclic} & CLIP & SwinB & 46.0 &- & 46.2 & - & - & - & -\\ 
    \midrule
    \textbf{VLDet-B} (Ours) & O365 & ViT-B & 54.2 & 50.4 & 51.2 & 16.9 & 30.1 & 43.8 & 35.6\\
    \textbf{VLDet-L} (Ours) & O365 & ViT-L & \textbf{58.7} & 53.9 & \textbf{55.2} & \textbf{24.8} & \textbf{44.7} & \textbf{48.0} & \textbf{44.6} \\
    \bottomrule
  \end{tabular}
  }
  
  \vspace{-5mm}
  \label{tab:ovd}
\end{table*}

\section{Experimental Results}
\label{sec:results}

\textbf{Datasets.} Following previous works \citep{zareian2021open, gu2021open, zhong2022regionclip, cheng2024yolo}, we evaluated our approach on two public datasets: COCO2017 \citep{lin2014microsoft} and LVIS \citep{gupta2019lvis}.
To evaluate the performance of open-vocabulary object detection, we follow the standard protocol to split the object categories into \textit{Base} and \textit{Novel} sets, train the model with only annotations of the \textit{Base} categories, and evaluate on \textit{Base}, \textit{Novel}, and \textit{All} all together. For COCO2017, We follow the data split of \cite{zareian2021open} with 48 \textit{Base} categories and 17 \textit{Novel} categories, and \textit{ALL} set contains 65 categories., which yields 107,761 training images and 4,836 test images. On LVIS, following \cite{gu2021open}, we adopt the category split with 866 \textit{Base} categories (common and frequent objects) and 337 \textit{Novel} categories (rare objects).
Additionally, VLDet is pretrained with only one public dataset, Objects365, with captions generated by BLIP-2 through prompt, "Describe this image in one sentence.", containing about 1.7M image-caption pairs.

\textbf{Implementation Details.} We adopted the two-stage Cascade RCNN \citep{cai2019cascade} as our basic detection framework. 
We employed ViT-B and ViT-L image encoders and the corresponding text encoders from CLIP as our image encoder and text encoder, respectively. 
During pre-training, we adopted the learning rate of $1\times10^{-5}$ for language backbone and $1\times10^{-4}$ for the rest modules.
We configured the AdamW optimizer with weight decay of $1\times10^{-4}$. Batch size is 2 for our VLDet-B and 1 for VLDet-L.
In the OVD fine-tuning stage, we reduced the learning rate to $1\times10^{-5}$ for all layers and froze the visual-to-language projection layers (V2L), while leaving other settings unchanged.
The model is trained for a maximum of 50 epochs during pre-training and 15 epochs during fine-tuning.
In addition, we set the maximum token length for captions to 64 and pad class names to the longest one (\emph{e.g.} 5 tokens for each class on COCO2017).
All experiments are conducted on 64 NVIDIA A100 GPUs.



\subsection{VLDet on Open-Vocabulary Object Detection}
We evaluate VLDet on COCO2017 and LVIS for OVD after pre-training on Objects365, the detailed comparison with baselines are shown in \cref{tab:ovd}. When experimenting on COCO2017 and LVIS, we train with only the $\textit{Base}$ categories (\textit{Common} and \textit{Frequent} for LVIS) and evaluate on different sets of categories.

\begin{figure}
  \centering
  \includegraphics[width=0.98\columnwidth]{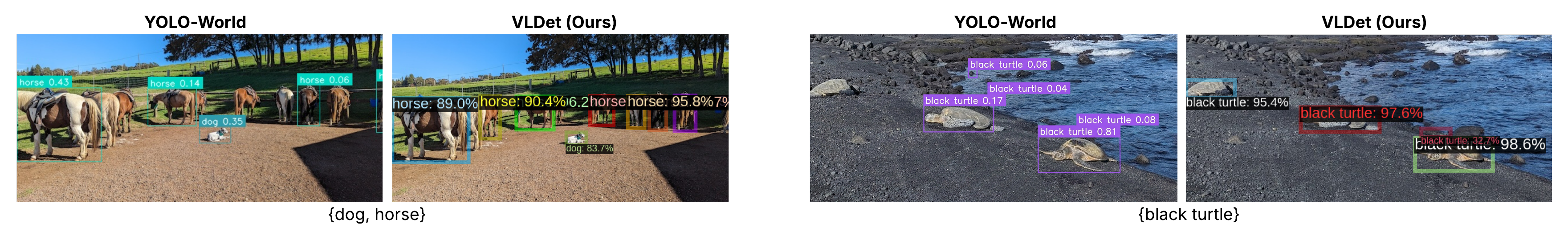}
  \caption{Comparison between YOLO-World and VLDet, VLDet detects objects (\emph{e.g.} "black turtle") more accurately.}
  \label{fig:result}
  \vspace{-6mm}
\end{figure}

Compared to all baselines, VLDet-L, utilizing the ViT-L backbone, achieves the highest AP across \textit{All} categories. More importantly, it also attains the best performance on \textit{Novel} categories, elevating the AP from 46.0 of CCKT-Det++~\citep{zhang2025cyclic} to 58.7 on COCO2017 and from 23.2 of BARON \citep{wu2023aligning} to 24.8 on LVIS. This demonstrates its extraordinary capability for generalization on new categories. Notably, for COCO2017, which contains 17 \textit{Novel} classes, the AP of our VLDet on \textit{Novel} categories surpasses that of \textit{Base} categories. Additionally, VLDet-B also achieves superior performance, further highlighting the benefits of our multi-level alignment loss for visual-language alignment. ViLD \citep{gu2021open} achieves the best AP on \textit{Base} classes with data augmentation through large-scale jittering \citep{ghiasi2021simple} and a significantly longer training schedule. Given that LVIS contains 337 \textit{Novel} categories, far more than COCO2017, VLDet-B with the ViT-B backbone performs limitedly. However, our larger model, VLDet-L outperforms all baselines across all metrics. Notably, when compared to YOLO-World-L \citep{cheng2024yolo}, which is pre-trained on a larger scale of data, our VLDet enhances the AP for \textit{Novel} categories by 4.43, demonstrating the superiority of the two-stage framework and SigRPN for OVD. We show the visualized comparison in \cref{fig:result}.

\subsection{VLDet on Closed-Set Object Detection}
For closed-set OD, we evaluated two scenarios: first, we conducted a zero-shot evaluation of our pre-trained model on the COCO2017 dataset (ZOD); second, we assessed the performance of our model after fine-tuning (FOD).

As shown in \cref{tab:od}, for ZOD, VLDet-L achieves superior performance with an AP of 45.8, surpassing baselines pre-trained on data of similar or even larger scale, such as YOLO-World \citep{cheng2024yolo}, which includes pseudo-labeled bounding boxes on CC3M and utilizes a vast category set extracted from grounding dataset captions. This demonstrates the efficacy of our framework, which adapts pre-aligned image and text encoders from CLIP with enriched visual-language semantic space and integrates a more robust visual-language alignment strategy. VLDet is more efficient, requiring 221 GLOPs for introduced layers versus 407 GLOPs for VLDyHead in GLIP \citep{li2022grounded}. 
Furthermore, for FOD, VLDet model also exhibits superior performance, achieving an AP of 55.9 after fine-tuning within 5 epochs. This surpasses not only open-world object detectors but also the state-of-the-art traditional object detector, YOLOv11 \citep{yolo11_ultralytics}.
We also include results from FIBER-B~\citep{dou2022coarse} and Grounding-DINO-L \citep{liu2023grounding}, which report great zero-shot performance. However, they are pre-trained on a dataset of significantly larger scale, even including COCO (thereby not strictly qualifying as zero-shot inference).

\begin{table}
  \centering
  \caption{Results for Zero-Shot Object Detection (ZOD) and Fine-Tuned Object Detection (FOD) on the COCO2017 dataset demonstrate that our VLDet model exhibits strong capabilities in both cases with pre-training solely on the Objects365 dataset. Numbers in \textcolor{gray}{gray} mean that they are not in zero-shot manner with pre-training data including COCO.}
  \begin{tabular}{@{}l l |c c @{}}
    \toprule
    \textbf{Method} & \textbf{Pre-train Data} & \makecell{ZOD\\(AP)} & \makecell{FOD\\(AP)} \\
    \midrule
    ViTDet-B~\citep{li2022exploring} & IN1K & - &  51.6\\
    YOLOv11-L~\citep{yolo11_ultralytics} & - & - &  53.4\\
    \midrule
    ViLD~\citep{gu2021open} & CLIP400M & 36.6 &  39.1\\
    RegionCLIP~\citep{zhong2022regionclip} & CC3M & 29.6 & -\\
    GLIP-T~\citep{li2022grounded} & O365+Gold & 44.6 & 53.8\\
    YOLO-World-M~\citep{cheng2024yolo} & O365+Gold & 42.8 & 51.2 \\
    YOLO-World-L~\citep{cheng2024yolo} & O365+Gold+CC & 45.1 & 53.3 \\
    \textcolor{gray}{FIBER-B}~\citep{dou2022coarse} & \textcolor{gray}{O365+COCO+CC+SBU+VG} & \textcolor{gray}{49.3} & \textcolor{gray}{58.4} \\
    \textcolor{gray}{Grounding-DINO-L}~\citep{liu2023grounding} & \textcolor{gray}{O365+COCO+OI+Gold+Cap4M+RefC} & \textcolor{gray}{60.7} & \textcolor{gray}{62.6} \\
    \midrule
    \textbf{VLDet-B} (Ours) & O365 & 43.7 & 54.1 \\
    \textbf{VLDet-L} (Ours) & O365 & \textbf{45.8} & \textbf{55.9}\\
    \bottomrule
  \end{tabular}
  
  \label{tab:od}
\end{table}

\subsection{Ablation Studies}

\begin{table}
  \centering
   \caption{$\mathcal{L}_{ICL}$ and $\mathcal{L}_{AAL}$ contribute to enhanced detection performance, with $\mathcal{L}_{AAL}$ providing greater improvements for rare and common categories.}
  \begin{tabular}{@{}c c c  |c c c c@{}}
    \toprule
    $\mathcal{L}_{RAL}$ & $\mathcal{L}_{ICL}$ & $\mathcal{L}_{AAL}$  & $\text{AP}_{r}$ & $\text{AP}_{c}$ & $\text{AP}_{f}$ & AP \\
    \midrule
    \cmark & \xmark & \xmark & 12.43 & 22.65 & 42.93 & 31.56\\
    \cmark & \cmark & \xmark & 14.90 & 26.18 & \textbf{43.87} & 33.76\\
    \cmark & \cmark & \cmark & \textbf{16.93} & \textbf{30.14} & 43.84 & \textbf{35.62}\\
    \bottomrule
  \end{tabular}
 
  \vspace{-7mm}
  \label{tab:loss}
\end{table}

\textbf{Multi-Level Alignment Loss.}
In \cref{tab:loss}, we show the fine-tuning results of VLDet-B on LVIS dataset for open-vocabulary object detection (OVD) with different losses. We can find that both of our introduced image-level loss $\mathcal{L}_{ICL}$ and anchor-text binary alignment loss $\mathcal{L}_{AAL}$ can enhance not only the overall AP on all classes, but also the AP on \textit{Novel} categories. These results demonstrate that the multi-level loss is beneficial for visual-language alignment. Specifically, the mini-batch image contrastive loss elevates the AP for $Novel$ categories of the LVIS dataset from 12.43 to 14.90 by 2.47, and then SigRPN further improves it to 16.93 by 2.03. In addition, the overall AP in all categories also improved from 31.56 to 35.62.

Importantly, with SigRPN and $\mathcal{L}_{AAL}$, we observe that the AP for frequent classes remains consistent, while the AP for rare and common classes significantly increases. Notably, our training data does not include the rare categories, highlighting the exceptional capability of our SigRPN in generalizing to \textit{Novel} categories by enabling the RPN to binary-differentiate background and object of any class with visual-language multi-modal embedding space.

We compare the performance of VLDet-B on standard object detection tasks using the COCO2017 dataset, trained from scratch with varying mini-batch sizes to balance image-wise visual-language alignment with final detection accuracy. We find that a mini-batch size of 8 yields the best performance, achieving an AP of 40.98. Detailed results are in \cref{sec:minibatch}.



\textbf{Specific Design of Our VL-Encoder.}
In this section, we detail the design of our VL-Encoder by experimenting with our VLDet-B for traditional object detection on the COCO2017 dataset from scratch. Initially, we replaced the pre-aligned text encoder for ViT-B/16 from CLIP with ViT-B/32, while retaining ViT-B/16 as the image encoder. We observed that the pre-aligned backbones, even when frozen, provide significantly higher AP, highlighting the superiority of inheriting enriched pre-aligned semantic information from CLIP.
Subsequently, we found that employing a multi-scale backbone with VL-PUB, as opposed to a single-scale model, improves the AP from 35.78 to 38.03. This demonstrates that pyramid features enhance detection accuracy by providing more spatial information. Furthermore, by comparing our model trained with and without captions, we discovered that incorporating captions alongside class prompts provides additional contextual information, enhancing the AP from 38.03 to 38.89. Besides, we found that adding $\mathcal{L}_{ICL}$ improves AP from 38.89 to 40.18 and the VL-Fuse layer further increases it to 40.98. Finally, we experimented with the text prompt template from previous work \citep{radford2021learning}, such as "A photo of \{label\}", but observed no performance gain. This approach also requires more tokens compared to using plain class names (\emph{i.e.} only 5 tokens on COCO), introducing additional computational cost. Therefore, we design our class prompts with plain class names.


\begin{table}
  \centering
  \caption{Results of different variants demonstrate the efficacy of our design.}
  \begin{tabular}{@{}l |c c c@{}}
    \toprule
    \textbf{Model} & AP & $\text{AP}_{50}$ & $\text{AP}_{75}$ \\
    \midrule
    Not pre-aligned Backbones & 34.86 & 55.96 &  36.78\\
    Pre-aligned, Frozen $E_L$ & 37.29 & 58.46 &  39.85\\
    Pre-aligned, Active $E_L$ & \textbf{38.03} & \textbf{59.29} &  \textbf{40.91}\\
    \midrule
    Single-scale Backbone & 35.78 & 53.13 &  38.6\\
    Multi-scale Backbone & \textbf{38.03} & \textbf{59.29} &  \textbf{40.91}\\
    \midrule
    Plain model (Only class prompt) & 38.03 & 59.29 &  40.91\\
    + Caption & 38.89 & 59.47& 41.76\\
    + Caption \& $\mathcal{L}_{ICL}$ & 40.18 & 62.07 & 43.22\\
    + Caption \& $\mathcal{L}_{ICL}$ \& VL-Fuse & \textbf{40.98} & \textbf{62.58}& \textbf{44.25}\\
    \midrule
    Class-template prompt (VLDet) & 40.88 & 62.68 & 44.33 \\
    Class-name prompt (VLDet) & 40.98 & 62.58 & 44.25 \\
    \bottomrule
  \end{tabular}
  \vspace{-2mm}
  \label{tab:ablation}
\end{table}

\textbf{Fine-tuning Strategies.}
For fine-tuning the pre-trained model on COCO2017 or LVIS for OVD, it is essential to freeze the visual projection layer (\emph{i.e.} V2L) \citep{zareian2021open}. Without this step, the model is prone to overfitting to \textit{Base} classes, resulting in a significant decline in performance on \textit{Novel} categories. As shown in \cref{tab:freeze}, when no layers are frozen, the AP on \textit{Base} categories (\emph{i.e.} $\text{AP}_{c}$ and $\text{AP}_{f}$) remains high, but $\text{AP}_{r}$ is notably low due to overfitting. We also observed that freezing the language background can lead to some loss in accuracy. The optimal fine-tuning strategy involves freezing the two V2L layers.

\begin{table}
  \centering
  \caption{Freeze different layers for fine-tuning on OVD. $\text{E}_L$ denote the language encoder, $\text{V2L}_1$ denote the final logit projection layer in SigRPN, and $\text{V2L}_2$ denote the logit projection layer in ROI.}
  \begin{tabular}{@{}c c c  |c c c c@{}}
    \toprule
    $\text{E}_L$ & $\text{V2L}_1$ & $\text{V2L}_2$ & AP & $\text{AP}_{r}$ & $\text{AP}_{c}$ & $\text{AP}_{f}$ \\
    \midrule
    \xmark & \xmark & \xmark & 44.03 & 14.77 & \textbf{45.44} & \textbf{48.04}\\
    \xmark & \xmark & \cmark &  44.09 & 24.24 &  43.85 & 47.88\\
    \xmark & \cmark & \cmark & \textbf{44.33}& \textbf{24.83} & 44.33  & 47.84\\
    \cmark & \xmark & \cmark  & 40.82	 & 24.65	& 37.17	& 46.97  \\
    \cmark & \cmark & \cmark & 41.78	&24.26	&38.89	&47.49\\
    \bottomrule
  \end{tabular}
  \label{tab:freeze}
  \vspace{-7mm}
\end{table}

\section{Conclusion}
We successfully extend the pre-aligned image and text encoders from CLIP, with its extensive visual-language latent space, into a novel OVD framework, Visual-Language Detection (VLDet). 
VLDet preserves the structure of feature pyramid, following the fusion of visual and language features, to boost spatial signals. 
We further introduce three constrastive losses to achieve the multi-level, fine-grained visual-language alignment, resulting superior detection performance on the OVD scenario.
VLDet significantly elevates the AP on novel categories by 27.6\% on COCO2017 and 6.9\% on LVIS, and outperforms zero-shot object detection baselines when pre-trained solely on Objects365.

\bibliography{iclr2026_conference}
\bibliographystyle{iclr2026_conference}

\appendix
\section*{Appendix}

\section{Bi-directional Cross-attention} \label{sec:crossattn}
Bi-directional Cross-attention is widely used for vision-language feature fusion \citep{li2022grounded, liu2023grounding}.
Specifically, the mechanism treats one modality as the Query and the other as the Key/Value. This occurs in two directions:
\begin{itemize}
    \item \textbf{Visual-to-Language}: The visual features attend to the language features to incorporate semantic context.

\begin{equation}
    \boldsymbol{v}_{out} = \boldsymbol{v} +\text{MHCA}( Q=\boldsymbol{v}W_q^{\boldsymbol{v}}, K=\boldsymbol{l}W_k^{\boldsymbol{l}}, V=\boldsymbol{l}W_v^{\boldsymbol{l}})
\end{equation} 

    \item \textbf{Language-to-Visual}: The language features attend to the visual features to ground the text in the specific image content.

\begin{equation}
    \boldsymbol{l}_{out} = \boldsymbol{l} +\text{MHCA}( Q=\boldsymbol{l}W_q^{\boldsymbol{l}}, K=\boldsymbol{v}W_k^{\boldsymbol{v}}, V=\boldsymbol{v}W_v^{\boldsymbol{v}})
\end{equation}
\end{itemize}

where $W$ terms are learnable projection matrices and MHCA denotes multi-head cross-attention. This mechanism is applied in \cref{eq:single} for single-scale features and \cref{eq:multi} for multi-scale features.

\section{Exploration on Mini-batch Size} \label{sec:minibatch}
While previous work, such as CLIP, suggests that large batch sizes work better for contrastive learning, they may dominate the training process and degrade the final detection performance.

\begin{figure}
  \centering
  \includegraphics[width=0.85\columnwidth]{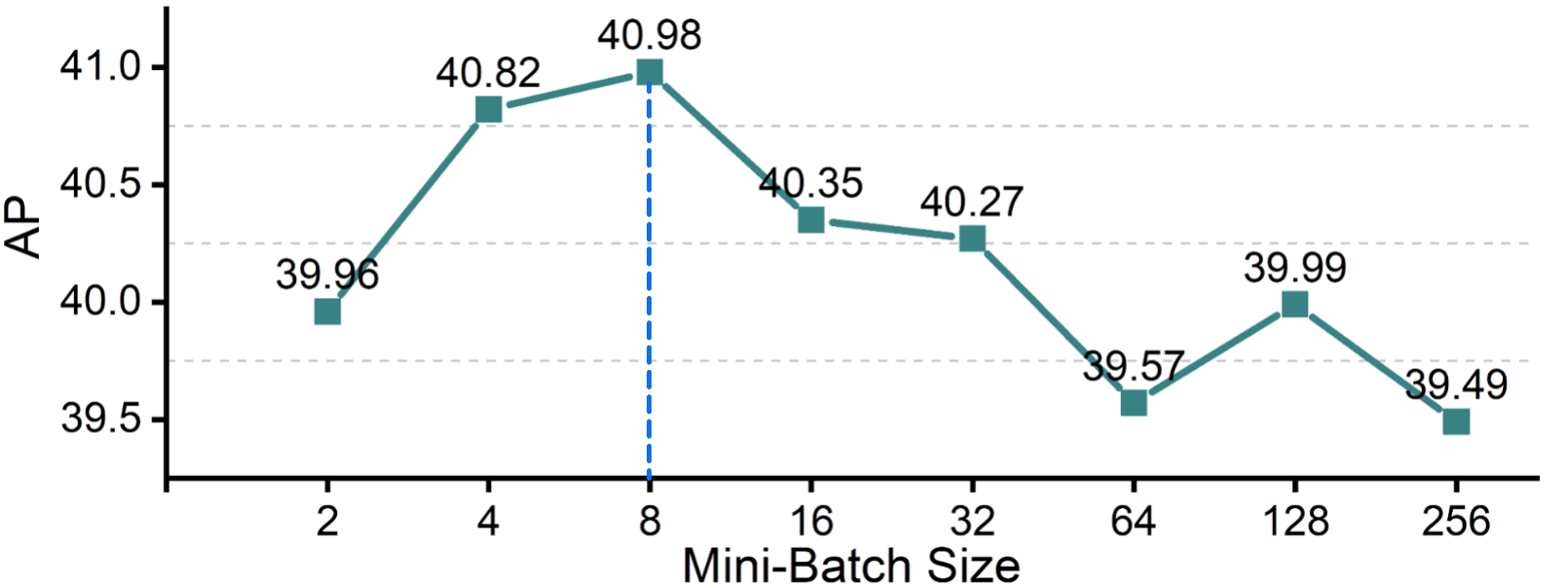}
  \caption{Image-level Contrastive Loss, $\mathcal{L}_{ICL}$ performs the best with mini-batch size of 8.}
  \label{fig:size}
\end{figure}

To balance image-wise and region-wise visual-language alignment, we explore a range of mini-batch sizes. Our experiments demonstrate that a batch size of 8 provides the best performance, as shown in \cref{fig:size}.

\section{Compared to Self-Distilled Baseline}
In the main text, our comparisons primarily focused on baselines that use ResNet as the vision backbone and rely on knowledge distillation to transfer visual-language (VL) knowledge from large VLMs like CLIP.

Here, we provide an additional, critical comparison with a state-of-the-art method that also directly leverages the Vision Transformer (ViT) from pre-aligned VLMs: CLIPSelf \citep{wu2023clipself}. CLIPSelf employs a self-distillation technique to refine the dense feature map of a CLIP ViT, making it more suitable for localization in dense prediction tasks. This approach enables the direct use of ViT backbones for open-vocabulary object detection (OVD).

As shown in \cref{tab:self-distill}, VLDet significantly outperforms CLIPSelf on COCO novel categories benchmarks with both ViT-B and ViT-L as vision backbone. This consistently superior performance further validates the effectiveness and robustness of our framework by adapting pre-aligned vision encoder into OVD detector through multi-level fine-grained contrastive loss.

\begin{table*}
  \centering
  \caption{VLDet outperforms self-distilled baseline for $Novel$ categories with various model sizes.}
  \resizebox{\textwidth}{!}{%
  \begin{tabular}{@{}l l |c c c | c c c c@{}}
    \toprule
    \multirow{2}{*}{\textbf{Method}} & \multirow{2}{*}{\textbf{Pre-train Data}} & \multicolumn{3}{c}{\textbf{ViT-B}} & \multicolumn{3}{|c}{\textbf{ViT-L}}\\
    & & $\text{AP}_{Novel}$ & $\text{AP}_{Base}$ & $\text{AP}_{All}$ & $\text{AP}_{Novel}$ & $\text{AP}_{Base}$ & $\text{AP}_{All}$\\
    \midrule
    F-ViT+CLIPSelf ~\citep{wu2023clipself} & CLIP  & 37.6 &- & - & 44.3 & - & - \\ 
    \textbf{VLDet} (Ours) & O365 & 54.2 & 50.4 & 51.2 & 58.7 & 53.9 & 55.2\\
    \bottomrule
  \end{tabular}
  }
  
  \label{tab:self-distill}
\end{table*}
\end{document}

%% file: math_commands.tex
\usepackage{amsmath,amsfonts,bm}

\def\eqref#1{equation~\ref{#1}}

\def\1{\bm{1}}

\DeclareMathAlphabet{\mathsfit}{\encodingdefault}{\sfdefault}{m}{sl}
\SetMathAlphabet{\mathsfit}{bold}{\encodingdefault}{\sfdefault}{bx}{n}



%% file: iclr2026_conference.bib
@inproceedings{zareian2021open,
  title={Open-vocabulary object detection using captions},
  author={Zareian, Alireza and Rosa, Kevin Dela and Hu, Derek Hao and Chang, Shih-Fu},
  booktitle={Proceedings of the IEEE/CVF Conference on Computer Vision and Pattern Recognition},
  pages={14393--14402},
  year={2021}
}

@inproceedings{lin2017feature,
  title={Feature pyramid networks for object detection},
  author={Lin, Tsung-Yi and Doll{\'a}r, Piotr and Girshick, Ross and He, Kaiming and Hariharan, Bharath and Belongie, Serge},
  booktitle={Proceedings of the IEEE conference on computer vision and pattern recognition},
  pages={2117--2125},
  year={2017}
}

@inproceedings{li2022exploring,
  title={Exploring plain vision transformer backbones for object detection},
  author={Li, Yanghao and Mao, Hanzi and Girshick, Ross and He, Kaiming},
  booktitle={European conference on computer vision},
  pages={280--296},
  year={2022},
  organization={Springer}
}

@inproceedings{li2022grounded,
  title={Grounded language-image pre-training},
  author={Li, Liunian Harold and Zhang, Pengchuan and Zhang, Haotian and Yang, Jianwei and Li, Chunyuan and Zhong, Yiwu and Wang, Lijuan and Yuan, Lu and Zhang, Lei and Hwang, Jenq-Neng and others},
  booktitle={Proceedings of the IEEE/CVF Conference on Computer Vision and Pattern Recognition},
  pages={10965--10975},
  year={2022}
}

@article{liu2023grounding,
  title={Grounding dino: Marrying dino with grounded pre-training for open-set object detection},
  author={Liu, Shilong and Zeng, Zhaoyang and Ren, Tianhe and Li, Feng and Zhang, Hao and Yang, Jie and Jiang, Qing and Li, Chunyuan and Yang, Jianwei and Su, Hang and others},
  journal={arXiv preprint arXiv:2303.05499},
  year={2023}
}

@inproceedings{cheng2024yolo,
  title={Yolo-world: Real-time open-vocabulary object detection},
  author={Cheng, Tianheng and Song, Lin and Ge, Yixiao and Liu, Wenyu and Wang, Xinggang and Shan, Ying},
  booktitle={Proceedings of the IEEE/CVF Conference on Computer Vision and Pattern Recognition},
  pages={16901--16911},
  year={2024}
}

@article{bica2024improving,
  title={Improving fine-grained understanding in image-text pre-training},
  author={Bica, Ioana and Ili{\'c}, Anastasija and Bauer, Matthias and Erdogan, Goker and Bo{\v{s}}njak, Matko and Kaplanis, Christos and Gritsenko, Alexey A and Minderer, Matthias and Blundell, Charles and Pascanu, Razvan and others},
  journal={arXiv preprint arXiv:2401.09865},
  year={2024}
}

@article{gu2021open,
  title={Open-vocabulary object detection via vision and language knowledge distillation},
  author={Gu, Xiuye and Lin, Tsung-Yi and Kuo, Weicheng and Cui, Yin},
  journal={arXiv preprint arXiv:2104.13921},
  year={2021}
}

@inproceedings{zhong2022regionclip,
  title={Regionclip: Region-based language-image pretraining},
  author={Zhong, Yiwu and Yang, Jianwei and Zhang, Pengchuan and Li, Chunyuan and Codella, Noel and Li, Liunian Harold and Zhou, Luowei and Dai, Xiyang and Yuan, Lu and Li, Yin and others},
  booktitle={Proceedings of the IEEE/CVF conference on computer vision and pattern recognition},
  pages={16793--16803},
  year={2022}
}

@inproceedings{radford2021learning,
  title={Learning transferable visual models from natural language supervision},
  author={Radford, Alec and Kim, Jong Wook and Hallacy, Chris and Ramesh, Aditya and Goh, Gabriel and Agarwal, Sandhini and Sastry, Girish and Askell, Amanda and Mishkin, Pamela and Clark, Jack and others},
  booktitle={International conference on machine learning},
  pages={8748--8763},
  year={2021},
  organization={PMLR}
}

@inproceedings{wu2023aligning,
  title={Aligning bag of regions for open-vocabulary object detection},
  author={Wu, Size and Zhang, Wenwei and Jin, Sheng and Liu, Wentao and Loy, Chen Change},
  booktitle={Proceedings of the IEEE/CVF conference on computer vision and pattern recognition},
  pages={15254--15264},
  year={2023}
}

@inproceedings{du2022learning,
  title={Learning to prompt for open-vocabulary object detection with vision-language model},
  author={Du, Yu and Wei, Fangyun and Zhang, Zihe and Shi, Miaojing and Gao, Yue and Li, Guoqi},
  booktitle={Proceedings of the IEEE/CVF Conference on Computer Vision and Pattern Recognition},
  pages={14084--14093},
  year={2022}
}

@article{kuo2022f,
  title={F-vlm: Open-vocabulary object detection upon frozen vision and language models},
  author={Kuo, Weicheng and Cui, Yin and Gu, Xiuye and Piergiovanni, AJ and Angelova, Anelia},
  journal={arXiv preprint arXiv:2209.15639},
  year={2022}
}

@software{yolo11_ultralytics,
  author = {Glenn Jocher and Jing Qiu},
  title = {Ultralytics YOLO11},
  version = {11.0.0},
  year = {2024},
  url = {https://github.com/ultralytics/ultralytics},
  orcid = {0000-0001-5950-6979, 0000-0002-7603-6750, 0000-0003-3783-7069},
  license = {AGPL-3.0}
}

@inproceedings{ghiasi2021simple,
  title={Simple copy-paste is a strong data augmentation method for instance segmentation},
  author={Ghiasi, Golnaz and Cui, Yin and Srinivas, Aravind and Qian, Rui and Lin, Tsung-Yi and Cubuk, Ekin D and Le, Quoc V and Zoph, Barret},
  booktitle={Proceedings of the IEEE/CVF conference on computer vision and pattern recognition},
  pages={2918--2928},
  year={2021}
}

@inproceedings{wu2018unsupervised,
  title={Unsupervised feature learning via non-parametric instance discrimination},
  author={Wu, Zhirong and Xiong, Yuanjun and Yu, Stella X and Lin, Dahua},
  booktitle={Proceedings of the IEEE conference on computer vision and pattern recognition},
  pages={3733--3742},
  year={2018}
}

@inproceedings{zhai2023sigmoid,
  title={Sigmoid loss for language image pre-training},
  author={Zhai, Xiaohua and Mustafa, Basil and Kolesnikov, Alexander and Beyer, Lucas},
  booktitle={Proceedings of the IEEE/CVF International Conference on Computer Vision},
  pages={11975--11986},
  year={2023}
}

@article{zeng2023x,
  title={X 2-vlm: All-in-one pre-trained model for vision-language tasks},
  author={Zeng, Yan and Zhang, Xinsong and Li, Hang and Wang, Jiawei and Zhang, Jipeng and Zhou, Wangchunshu},
  journal={IEEE Transactions on Pattern Analysis and Machine Intelligence},
  year={2023},
  publisher={IEEE}
}

@article{yao2022detclip,
  title={Detclip: Dictionary-enriched visual-concept paralleled pre-training for open-world detection},
  author={Yao, Lewei and Han, Jianhua and Wen, Youpeng and Liang, Xiaodan and Xu, Dan and Zhang, Wei and Li, Zhenguo and Xu, Chunjing and Xu, Hang},
  journal={Advances in Neural Information Processing Systems},
  volume={35},
  pages={9125--9138},
  year={2022}
}

@inproceedings{redmon2016you,
  title={You only look once: Unified, real-time object detection},
  author={Redmon, J},
  booktitle={Proceedings of the IEEE conference on computer vision and pattern recognition},
  year={2016}
}

@inproceedings{redmon2017yolo9000,
  title={YOLO9000: better, faster, stronger},
  author={Redmon, Joseph and Farhadi, Ali},
  booktitle={Proceedings of the IEEE conference on computer vision and pattern recognition},
  pages={7263--7271},
  year={2017}
}

@article{redmon2018yolov3,
  title={Yolov3: An incremental improvement},
  author={Redmon, Joseph},
  journal={arXiv preprint arXiv:1804.02767},
  year={2018}
}

@inproceedings{carion2020end,
  title={End-to-end object detection with transformers},
  author={Carion, Nicolas and Massa, Francisco and Synnaeve, Gabriel and Usunier, Nicolas and Kirillov, Alexander and Zagoruyko, Sergey},
  booktitle={European conference on computer vision},
  pages={213--229},
  year={2020},
  organization={Springer}
}

@article{zhang2022dino,
  title={Dino: Detr with improved denoising anchor boxes for end-to-end object detection},
  author={Zhang, Hao and Li, Feng and Liu, Shilong and Zhang, Lei and Su, Hang and Zhu, Jun and Ni, Lionel M and Shum, Heung-Yeung},
  journal={arXiv preprint arXiv:2203.03605},
  year={2022}
}

@article{ren2016faster,
  title={Faster R-CNN: Towards real-time object detection with region proposal networks},
  author={Ren, Shaoqing and He, Kaiming and Girshick, Ross and Sun, Jian},
  journal={IEEE transactions on pattern analysis and machine intelligence},
  volume={39},
  number={6},
  pages={1137--1149},
  year={2016},
  publisher={IEEE}
}

@inproceedings{he2017mask,
  title={Mask r-cnn},
  author={He, Kaiming and Gkioxari, Georgia and Doll{\'a}r, Piotr and Girshick, Ross},
  booktitle={Proceedings of the IEEE international conference on computer vision},
  pages={2961--2969},
  year={2017}
}

@article{cai2019cascade,
  title={Cascade R-CNN: High quality object detection and instance segmentation},
  author={Cai, Zhaowei and Vasconcelos, Nuno},
  journal={IEEE transactions on pattern analysis and machine intelligence},
  volume={43},
  number={5},
  pages={1483--1498},
  year={2019},
  publisher={IEEE}
}

@inproceedings{shao2019objects365,
  title={Objects365: A large-scale, high-quality dataset for object detection},
  author={Shao, Shuai and Li, Zeming and Zhang, Tianyuan and Peng, Chao and Yu, Gang and Zhang, Xiangyu and Li, Jing and Sun, Jian},
  booktitle={Proceedings of the IEEE/CVF international conference on computer vision},
  pages={8430--8439},
  year={2019}
}

@inproceedings{plummer2015flickr30k,
  title={Flickr30k entities: Collecting region-to-phrase correspondences for richer image-to-sentence models},
  author={Plummer, Bryan A and Wang, Liwei and Cervantes, Chris M and Caicedo, Juan C and Hockenmaier, Julia and Lazebnik, Svetlana},
  booktitle={Proceedings of the IEEE international conference on computer vision},
  pages={2641--2649},
  year={2015}
}

@inproceedings{kamath2021mdetr,
  title={Mdetr-modulated detection for end-to-end multi-modal understanding},
  author={Kamath, Aishwarya and Singh, Mannat and LeCun, Yann and Synnaeve, Gabriel and Misra, Ishan and Carion, Nicolas},
  booktitle={Proceedings of the IEEE/CVF international conference on computer vision},
  pages={1780--1790},
  year={2021}
}

@inproceedings{hudson2019gqa,
  title={Gqa: A new dataset for real-world visual reasoning and compositional question answering},
  author={Hudson, Drew A and Manning, Christopher D},
  booktitle={Proceedings of the IEEE/CVF conference on computer vision and pattern recognition},
  pages={6700--6709},
  year={2019}
}

@article{krishna2017visual,
  title={Visual genome: Connecting language and vision using crowdsourced dense image annotations},
  author={Krishna, Ranjay and Zhu, Yuke and Groth, Oliver and Johnson, Justin and Hata, Kenji and Kravitz, Joshua and Chen, Stephanie and Kalantidis, Yannis and Li, Li-Jia and Shamma, David A and others},
  journal={International journal of computer vision},
  volume={123},
  pages={32--73},
  year={2017},
  publisher={Springer}
}

@inproceedings{sharma2018conceptual,
  title={Conceptual captions: A cleaned, hypernymed, image alt-text dataset for automatic image captioning},
  author={Sharma, Piyush and Ding, Nan and Goodman, Sebastian and Soricut, Radu},
  booktitle={Proceedings of the 56th Annual Meeting of the Association for Computational Linguistics (Volume 1: Long Papers)},
  pages={2556--2565},
  year={2018}
}

@article{ordonez2011im2text,
  title={Im2text: Describing images using 1 million captioned photographs},
  author={Ordonez, Vicente and Kulkarni, Girish and Berg, Tamara},
  journal={Advances in neural information processing systems},
  volume={24},
  year={2011}
}

@inproceedings{lin2014microsoft,
  title={Microsoft coco: Common objects in context},
  author={Lin, Tsung-Yi and Maire, Michael and Belongie, Serge and Hays, James and Perona, Pietro and Ramanan, Deva and Doll{\'a}r, Piotr and Zitnick, C Lawrence},
  booktitle={Computer Vision--ECCV 2014: 13th European Conference, Zurich, Switzerland, September 6-12, 2014, Proceedings, Part V 13},
  pages={740--755},
  year={2014},
  organization={Springer}
}

@inproceedings{gupta2019lvis,
  title={Lvis: A dataset for large vocabulary instance segmentation},
  author={Gupta, Agrim and Dollar, Piotr and Girshick, Ross},
  booktitle={Proceedings of the IEEE/CVF conference on computer vision and pattern recognition},
  pages={5356--5364},
  year={2019}
}

@inproceedings{zong2023detrs,
  title={Detrs with collaborative hybrid assignments training},
  author={Zong, Zhuofan and Song, Guanglu and Liu, Yu},
  booktitle={Proceedings of the IEEE/CVF international conference on computer vision},
  pages={6748--6758},
  year={2023}
}

@article{ren2023strong,
  title={A strong and reproducible object detector with only public datasets},
  author={Ren, Tianhe and Yang, Jianwei and Liu, Shilong and Zeng, Ailing and Li, Feng and Zhang, Hao and Li, Hongyang and Zeng, Zhaoyang and Zhang, Lei},
  journal={arXiv preprint arXiv:2304.13027},
  year={2023}
}

@inproceedings{zhou2022detecting,
  title={Detecting twenty-thousand classes using image-level supervision},
  author={Zhou, Xingyi and Girdhar, Rohit and Joulin, Armand and Kr{\"a}henb{\"u}hl, Philipp and Misra, Ishan},
  booktitle={European Conference on Computer Vision},
  pages={350--368},
  year={2022},
  organization={Springer}
}

@inproceedings{gao2022open,
  title={Open vocabulary object detection with pseudo bounding-box labels},
  author={Gao, Mingfei and Xing, Chen and Niebles, Juan Carlos and Li, Junnan and Xu, Ran and Liu, Wenhao and Xiong, Caiming},
  booktitle={European Conference on Computer Vision},
  pages={266--282},
  year={2022},
  organization={Springer}
}

@inproceedings{shen2024aligning,
  title={Aligning and prompting everything all at once for universal visual perception},
  author={Shen, Yunhang and Fu, Chaoyou and Chen, Peixian and Zhang, Mengdan and Li, Ke and Sun, Xing and Wu, Yunsheng and Lin, Shaohui and Ji, Rongrong},
  booktitle={Proceedings of the IEEE/CVF Conference on Computer Vision and Pattern Recognition},
  pages={13193--13203},
  year={2024}
}

@inproceedings{jiang2024t,
  title={T-rex2: Towards generic object detection via text-visual prompt synergy},
  author={Jiang, Qing and Li, Feng and Zeng, Zhaoyang and Ren, Tianhe and Liu, Shilong and Zhang, Lei},
  booktitle={European Conference on Computer Vision},
  pages={38--57},
  year={2024},
  organization={Springer}
}

@inproceedings{wang2025ov,
  title={Ov-dquo: Open-vocabulary detr with denoising text query training and open-world unknown objects supervision},
  author={Wang, Junjie and Chen, Bin and Kang, Bin and Li, Yulin and Xian, Weizhi and Chen, Yichi and Xu, Yong},
  booktitle={Proceedings of the AAAI Conference on Artificial Intelligence},
  volume={39},
  number={7},
  pages={7762--7770},
  year={2025}
}

@article{zhang2025cyclic,
  title={Cyclic Contrastive Knowledge Transfer for Open-Vocabulary Object Detection},
  author={Zhang, Chuhan and Zhu, Chaoyang and Dong, Pingcheng and Chen, Long and Zhang, Dong},
  journal={arXiv preprint arXiv:2503.11005},
  year={2025}
}

@article{wu2023clipself,
  title={Clipself: Vision transformer distills itself for open-vocabulary dense prediction},
  author={Wu, Size and Zhang, Wenwei and Xu, Lumin and Jin, Sheng and Li, Xiangtai and Liu, Wentao and Loy, Chen Change},
  journal={arXiv preprint arXiv:2310.01403},
  year={2023}
}

@article{dou2022coarse,
  title={Coarse-to-fine vision-language pre-training with fusion in the backbone},
  author={Dou, Zi-Yi and Kamath, Aishwarya and Gan, Zhe and Zhang, Pengchuan and Wang, Jianfeng and Li, Linjie and Liu, Zicheng and Liu, Ce and LeCun, Yann and Peng, Nanyun and others},
  journal={Advances in neural information processing systems},
  volume={35},
  pages={32942--32956},
  year={2022}
}

@inproceedings{zhang2024transformer,
  title={Transformer-based selective super-resolution for efficient image refinement},
  author={Zhang, Tianyi and Kasichainula, Kishore and Zhuo, Yaoxin and Li, Baoxin and Seo, Jae-Sun and Cao, Yu},
  booktitle={Proceedings of the AAAI Conference on Artificial Intelligence},
  volume={38},
  number={7},
  pages={7305--7313},
  year={2024}
}

@inproceedings{zhang2024patch,
  title={Patch-based selection and refinement for early object detection},
  author={Zhang, Tianyi and Kasichainula, Kishore and Zhuo, Yaoxin and Li, Baoxin and Seo, Jae-Sun and Cao, Yu},
  booktitle={Proceedings of the IEEE/CVF Winter Conference on Applications of Computer Vision},
  pages={729--738},
  year={2024}
}

@article{zhang2024context,
  title={Context-Aware Token Selection and Packing for Enhanced Vision Transformer},
  author={Zhang, Tianyi and Li, Baoxin and Seo, Jae-sun and Cao, Yu},
  journal={arXiv preprint arXiv:2410.23608},
  year={2024}
}

@article{cai2025does,
  title={Does Tone Change the Answer? Evaluating Prompt Politeness Effects on Modern LLMs: GPT, Gemini, LLaMA},
  author={Cai, Hanyu and Shen, Binqi and Jin, Lier and Hu, Lan and Fan, Xiaojing},
  journal={arXiv preprint arXiv:2512.12812},
  year={2025}
}
